\documentclass[PHME, 2024]{PHMSociety}
\usepackage{graphicx}
\usepackage{amsmath}
\usepackage{booktabs}
\usepackage{pifont}
\usepackage{amsfonts}
\usepackage{amssymb}
\usepackage{dsfont} % indicator function

\newcommand{\cmark}{\ding{51}}%
\newcommand{\xmark}{\ding{55}}%

\begin{document}

% Paper Title
\title{Graph Neural Networks for Electric and Hydraulic Data Fusion to Enhance Short-term Forecasting of Pumped-storage Hydroelectricity }

% Authors List
\author{%			
	Raffael Theiler\authorNumber{1}, Olga Fink\authorNumber{2}
}

% Author Affiliations
\address{% This is a tabular environment so each affiliation needs to be separated by "\\" or "\tabularnewline"
	\affiliation{{1,2}}{EPFL, Lausanne, Vaud, 1015, Switzerland}{ %add emails
		{\email{raffael.theiler@epfl.ch}}\\ 
		{\email{olga.fink@epfl.ch}}
		} % emails input
}

% Create the title
\maketitle
\pagestyle{fancy}
\thispagestyle{plain}

\phmLicenseFootnote{Raffael Theiler}

% Abstract
\begin{abstract}

Pumped-storage hydropower plants (PSH) actively participate in grid power-frequency control and therefore often operate under dynamic conditions, which results in rapidly varying system states.
Predicting these dynamically changing states is essential for comprehending the underlying sensor and machine conditions. This understanding aids in  detecting anomalies and faults, ensuring the reliable operation of the connected power grid, and in identifying faulty and miscalibrated sensors.
PSH are complex, highly interconnected systems encompassing electrical and hydraulic subsystems, each characterized by their respective underlying networks that can individually be represented as graph.
To take advantage of this relational inductive bias, \textit{graph neural networks} (GNNs) have been separately applied to state forecasting tasks in the individual subsystems, but without considering their interdependencies.
In PSH, however, these subsystems depend on the same control input, making their operations highly interdependent and interconnected. 
Consequently, hydraulic and electrical sensor data should be fused across PSH subsystems to improve state forecasting accuracy.
This approach has not been explored in GNN literature yet because many available PSH graphs are limited to their respective subsystem boundaries, which makes the method unsuitable to be applied directly.
In this work, we introduce the application of  \textit{spectral-temporal graph neural networks}, which  leverage  self-attention mechanisms to  concurrently capture and learn meaningful subsystem interdependencies and the
dynamic patterns observed in electric and hydraulic sensors.
Our method effectively fuses data from the PSH's subsystems by operating on a unified, system-wide graph, learned directly from the data, This approach leads to demonstrably improved state forecasting performance and enhanced generalizability.
\end{abstract}

\section{Introduction}
In power grids, pumped-storage hydropower plants (PSH) are well-established
for large-scale energy storage due to their efficiency, scalablilty and flexibility.
In this role, these plants dynamically respond to potentially large fluctuations in grid demand.
In the transition towards smart grids, PSH sensor data 
is collected via \textit{wide area measurement systems (WAMS)} \cite{pagnierEmbeddingPowerFlow2021} and is stored
in centralized \textit{energy management systems (EMS)}.
By processing the aggregated WAMS data, 
modern EMS provide crucial  functionalities for PSH operators
such as load forecasting, 
real-time monitoring, 
distribution and demand-side management, 
and various decision support tools aimed at  
increasing efficiency and sustainability.
To enhance system reliability, EMS implement anomaly and sensor fault detection 
based on short-term forecasting and state estimation,
playing a pivotal role in preventing failures 
that could lead to widespread power grid outages and significant economic losses.
However, the dynamic operation of the PSH 
and the vast amounts of data transmitted by the WAMS 
significantly complicate the task. 
In the PSH environment, conventional state estimation 
is often ineffective
because the computation  can take several minutes \cite{liDynamicGraphBasedAnomaly2022}.
This delay leads to a rapid divergence 
between the most recent and the previously used system state
 for the estimation, 
resulting in numerous  false-positives when applied to anomaly detection. Consequently, it becomes challenging to maintain a comprehensive overview 
of the system's health and performance.
As a solution, deep-learning-based short-term state forecasting has recently been applied,
which offers significantly faster processing times 
and holds the potential to benefit from the additional data 
increasingly collected at a high sampling rate \cite{kundacinaStateEstimationElectric2022}.

Developing deep-learning-based state forecasting
for PSH is particularly challenging.
These challenges stem from the necessity to accurately represent
two distinct physical domains within the PSH: the hydraulic and electrical systems.
Although these domains are mechanically interconnected by electromagnetic generators, they are traditionally modeled independently in mechanical engineering.
This division mainly stems from the distinct dynamics governing each subsystem.
It is, therefore, challenging to model hydraulic and electrical domains simultaneously. 
Nonetheless, considering  the direct causal relationship between the systems -- wherein kinetic energy is transformed to electric energy --
we hypothesize that fusing data from both subsystems, which operate under a unified  control input, can significantly enhance state forecasting.

To address the challenge of fusing electric and hydraulic data, 
we posit  that both subsystems of the PSH consist of extensive networks, 
which are coarsely monitored with sensors  
that can be represented in the non-Euclidean graph domain. 
By operating on this more effective graph representation,
which can capture biases given by the PSH system architecture
and homophily biases,
addressing the phenomenon that sensor measurements tend to be connected with “similar” or “alike” others \cite{maHomophilyNecessityGraph2023},
\textit{graph neural networks} (GNNs) have recently gained significant attention.
When a graph is available, GNNs have been effectively applied in key applications 
to (hydro) power plants  \cite{liaoReviewGraphNeural2022}
and in the broader power grid environment. 
However, these methods depend on the availability of apriori graphs, 
derived from PSH's electrical and hydraulic network diagrams.
Therefore, their applicability is limited by the fact that,
although the underlying network structure 
of both electric and hydraulic subsystems of a PSH can be modeled as a graph, 
network diagrams for PSH exist typically only separately for each subsystem.
Consequently, most graph-based methods are confined 
within the boundaries of  their respective systems. 
To overcome this limitation, 
we propose learning a PSH sensor graph 
from latent dependencies in the data.
While it has previously been demonstrated that graph structures 
can be efficiently learned from data, 
this approach remains  unexplored  in the context of hydropower plants.
In light of this, inspired by \cite{caoSpectralTemporalGraph2021},
we propose using \textit{spectral-temporal graph neural networks} (STGNN) to learn a latent correlation graph structure across the entire PSH asset 
for  the fusion of electric and hydraulic data, leveraging the self-attention mechanism.

Compared to numerical simulation, our data-driven GNN-based methodology is computationally inexpensive
and does not require expert knowledge while maintaining interpretability, due to the accessibility of the learned graph.
Our proposed approach can be easily transferred to different PSH assets without any calibration.
To the best of our knowledge, there is no other work 
on data-driven electric and hydraulic data fusion for PSH using graph neural networks.

To summarize, in this work, we introduce the application of attention-based graph neural networks to effectively learn intra- and interdependencies 
 between the subsystems' sensors to enhance the short-term state forecast in the pumped 
storage power plant (PSH) environment. 
We tackle several challenges when applying  state forecasting to PSH:

\begin{itemize}
    \item In line with the dynamic operation of the PSH,
    the  dynamic behavior of sensors
    adds complexity to the forecasting. 
    We propose a \textit{spectral-temporal graph neural network} (STGNN)
    that effectively captures these patterns by incorporating the PSH' 
    underlying structural and homophily biases, such as load patterns that are reflected across sensor measurement sites.
    \item State forecasting in the PSH environ\-ment 
    de\-pends strongly on environ\-mental parameters, 
    such as temperature, daily load profiles, 
    and power grid customer-related factors, 
    which cannot be modeled in numerical simulations
    \cite{linSpatialTemporalResidentialShortTerm2021}.
    In contrast, our STGNN is able to learn these factors from data.
    \item The PSH is a spatially distributed complex system
    that spans across the hydraulic and  electric domains.
    We propose a graph learning module that learns a unified graph representation
    across the hydraulic and electrical subsystems from latent dependencies in the data.
    \item We assess the performance of our method on a multivariate PSH dataset containing 58 signals, showcasing the dynamic operation of the asset.
\end{itemize}

The reminder of this paper is organized as follows: 
Sec. \ref{sec:background_and_related_work} reviews relevant literature that focuses on graph-based deep learning and data fusion. Sec. \ref{sec:methodology}
introduces our STGNN approach. 
In Sec.~\ref{sec:case_study}, we discuss the case study conducted on a Swiss PSH plant, including the experimental and training setups. In Sec.~\ref{sec:results}, we present our results. 
Finally, Sec.~\ref{sec:conclusions} concludes this work and outlines future steps.

\section{Background and Related Work}
\label{sec:background_and_related_work}

Conventional machine learning applied to power systems 
have primarily  focussed on linear regression models and 
recurrent neural networks~\cite{zhengElectricLoadForecasting2017}. These methodologies continue to be effective and provide competitive results, particularly in areas like  
short-term load forecasting~\cite{guoMachineLearningBasedMethods2021}
and daily peak-energy demand forecasting
\cite{kimDailyPeakElectricityDemandForecasting2022}.
%GNN
Since their introduction, GNNs ~\cite{bronsteinGeometricDeepLearning2017} 
have been applied to many tasks in power systems.
By now, graph-based deep learning has become 
a well-established method for analyzing power system data,
thanks to its ability to include structural and homophily biases that cannot be 
modeled conventionally.
Mes\-sage-passing GNNs  have been successfully applied to state estimation 
\cite{kundacinaStateEstimationElectric2022},
and power flow estimations
\cite{ringsquandlPowerRelationalInductive2021}.
The same tasks have also been addressed using 
\textit{graph convolutional neural networks} (GCN)
\cite{fatahIntegratingPowerGrid2021}. 
In the broader power-grid environment,
GNNs are also used for wind speed forecasting in renewable energy \cite{liaoFaultDiagnosisPower2021a}.
% Downstream tasks
Additionally, GNN-based state forecasting was used in a range of downstream tasks in several previous research studies, including 
graph-based early fault detection for IIoT systems
\cite{zhaoDyEdgeGATDynamicEdge2024}, 
ano\-maly detection in the electrical grid  \cite{liDynamicGraphBasedAnomaly2022},
fault diagnosis for three-phase flow facility
\cite{chenInteractionAwareGraphNeural2021d}, predicting  dynamical grid stability
\cite{nauckPredictingBasinStability2022},
and physics-informed parameter and state estimations 
\cite{pagnierPhysicsInformedGraphicalNeural2021, pagnierEmbeddingPowerFlow2021}. 

% spatial-temporal
Other works have used spatial-temporal  extensions of GNNs
in the electrical domain
for residential load-forecasting \cite{linSpatialTemporalResidentialShortTerm2021}, fault diagnostics in power distribution systems \cite{nguyenSpatialTemporalRecurrentGraph2022}, 
and with complex-value extensions \cite{wuComplexValueSpatiotemporalGraph2022} for state forecasting.
In another lie of research, 
\cite{wangDynamicSpatiotemporalCorrelation2022} propose spatial-temporal graph learning for power flow analysis, where the graph is dynamically created from thresholded normalized mutual information.
% attention based
At the component feature level, attention-based graph learning (GAT) has been applied to power flow analysis
\cite{jeddiPhysicsInformedGraphAttentionbased2021} 
and in a different work for probabilistic power flow to quantify uncertainties of distribution power systems \cite{wuGraphAttentionEnabled2022}.
To the best of our knowledge, 
the approach of employing a self-attention mechanism 
at the graph level to learn a graph structure that
integrates electrical and hydraulic data using \textit{graph neural networks}
has not yet been addressed in previous research.

% data fusion
In the context of power systems, data fusion represents a crucial technique for enhancing the accuracy and reliability of forecasting algorithms by integrating diverse data sources.
In the electrical domain, the fusion of diverse electrical system information 
was utilized to estimate the voltage in distribution networks
\cite{zhuCrossDomainDataFusion2020}, using cross-correlations between individual transformers.
Another research study achieved state-of-the-art
multi-site photovoltaic (PV) power forecasting 
\cite{simeunovicSpatioTemporalGraphNeural2022},
fusing spatially distributed PV data by 
exploiting the intuition that PV systems provide a dense network of virtual weather stations.
Integrating weather station data was also explored for anomaly detection 
for the industrial internet of things 
\cite{wuGraphNeuralNetworks2022}.
Given its importance, modeling the interaction  between the PSH subsystems 
has previously been explored using higher-order numerical simulation (SIMSEN) \cite{simondSimulationToolLinking2006}.
However, operating numerical simulators in practice
requires precise calibration 
and, consequently, extensive documentation of the components, which is typically not readily available. 
This calibration step is indispensable due to the components exhibiting highly non-linear characteristics 
\cite{nicoletHighOrderModelingHydraulic2007}. 
Given the significant variations in designs across different PSH assets, and the necessity for 
expert knowledge (which is often unavailable), applying this simulator-based methodology  is often infeasible in real-world applications. 
For PSH, these limitations shift the focus to data-driven interaction modelling with GNNs, which is computationally affordable and does not necessitate expert knowledge, yet remains unexplored.

\section{Methodology}
\label{sec:methodology}
\begin{figure*}[tph]
    \centering
    \includegraphics[width=0.6\linewidth]{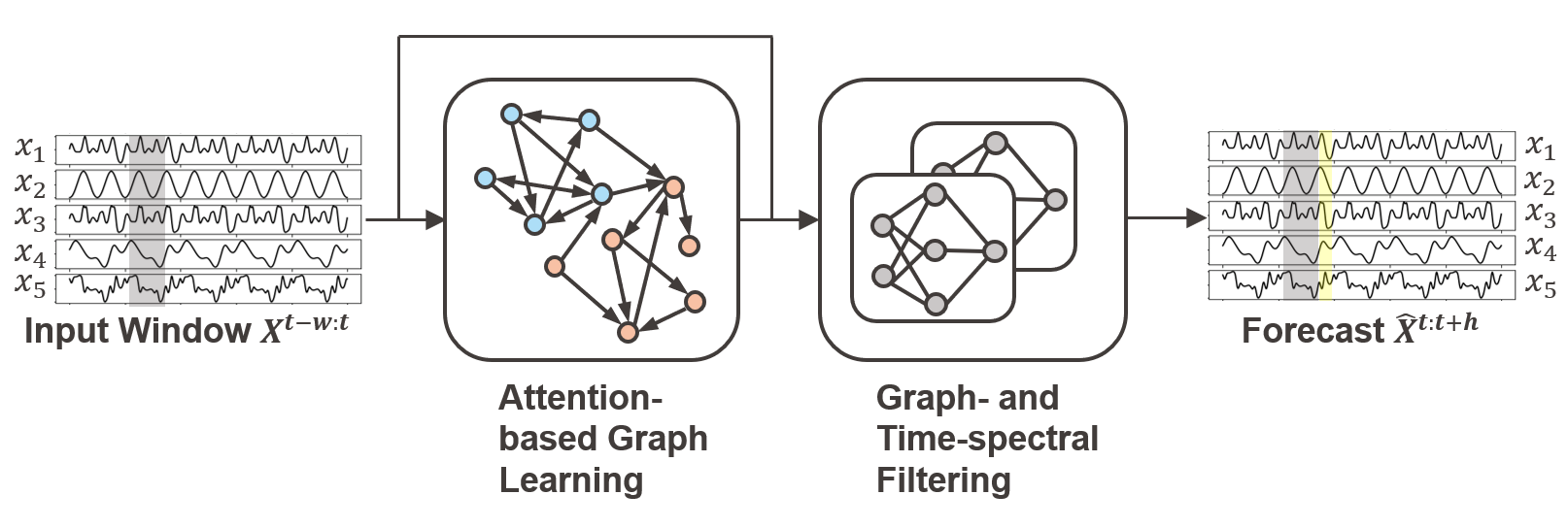}

  \caption{
  An overview of the two processing steps of our \textit{spectral-temporal graph neural network} to fuse data from the electrical and hydraulic subsystems of a PSH. Utilizing \textit{attention-based graph learning}, our method dynamically constructs a graph based on an input window \(\bar{\mathbf{X}}^{t-w:t}\). Subsequently, by operating on this graph, the \textit{graph- and time-spectral filtering} module efficiently extracts information from the hydraulic and electrical sensor data to forecast the subset of electrical sensors \(\widetilde{\mathbf{X}}_{\text{elec}}^{t:t+h}\). }
    \label{fig:methodology_overview}
\end{figure*}

This section introduces the Spectral-Temporal Graph Neural Network (STGNN) that we propose for the fusion of electric and hydraulic data  in PSH state forecasting. 
In Section \ref{sec:problem_formulation}, we define  the forecasting problem.
From  Section \ref{sec:node_features} onwards, 
we decompose the forecasting problem into learning 
the underlying latent graph structure from time series data (Sec. \ref{sec:graph_learning}).
Subsequently, on the learned graph, we introduce
graph-spectral and time-spectral filtering (Sec. \ref{sec:filtering_graph_and_time}).
An overview of the methodology is provided in Figure 
\ref{fig:methodology_overview}.

{\small
\textbf{Notation}: In this work, we use slicing notation denoted by the colon (\(:\)) symbol.
Given a matrix \( A \in \mathbb{R}^{m \times n} \), where \( m \) and \( n \) represent the number of rows and columns respectively, slicing is expressed as \(A[i:j, k:l]\) or \(A^{i:j, k:l}\).
This notation represents the selection of rows \( i \) through \( j-1 \) and columns \( k \) through \( l-1 \) of matrix \( A \). If \( i \) or \( k \) is omitted, it implies  starting from the first row or column, respectively. Similarly, if \( j \) or \( l \) is omitted, it implies selection until the last row or column, respectively.
We use $\otimes$ to denote element wise multiplication and
$\oplus$ for concatenation. The Frobenius norm is denoted as
\(\Vert \bullet \Vert_{\text{F}}\) .

}

\subsection{Problem Formulation}
\label{sec:problem_formulation}

The specific objective of this work is to provide accurate state forecasting for the electrical subsystem of the PSH.
Our approach utilizes learnable graph-spectral and time-spectral filtering to compute
state predictions (the forecast).
Let \( \bar{\mathbf{X}} \in \mathbb{R}^{T \times H + E} \) 
represent the smoothed time series data
\(\bar{\mathbf{X}} = S(\mathbf{X})\) 
computed from unsmoothed time series \(\mathbf{X}\) using the smoothing function $S$
for \( E \) sensors 
in the electric and \(H\) sensors in the hydraulic subsystem, respectively, over a time period \( T \). 
We define our forecasting model as a function \( M: \mathbb{R}^{w \times H+E} \rightarrow  \mathbb{R}^{h \times E} \)  for a specific point in time $w < t < T$,
that operates on input windows of the data of length $w$ sliced as
\( \bar{\mathbf{X}}[t-w:t] \).
The goal is to forecast the subset of electrical sensors
\(\bar{\mathbf{X}}[t:t+h, :E] \) 
for a horizon of size $h$. 
Model \(M\) operates on a graph \(\mathcal{G}\) that is either inferred from the input data by a parameterized function \(\mathcal{G}_\phi(\bar{\mathbf{X}}[t-w:t])\),
which is trained alongside $M$, or may be provided apriori.
We introduce two sets of learnable parameters: \(\theta\) for the filtering $M_\theta$ and \(\phi\) for the graph learning $\mathcal{G}_\phi$.
Thus, state forecasting and state reconstruction estimates denoted by \(\widetilde{\bullet}\) at a selected timepoint $t$, are computed by the model \(M\) as follows:
\begin{equation}
    \widetilde{\mathbf{X}}_{\text{elec}}^{t:t+h}, \widetilde{\mathbf{X}}_{\text{elec}}^{t-w:t} = M_\theta(\bar{\mathbf{X}}^{t-w:t} \mid \mathcal{G}_\phi)
\end{equation}

\subsection{Training Objective}

By employing a sliding window approach, we construct a training dataset $\mathcal{X}^{\text{Train}}$ of length \(N_\text{train}\), where \(N_\text{train}\) depends on the training data split.
We also construct analogous validation and test datasets $\mathcal{X}^{\text{Val}}$ and $\mathcal{X}^{\text{Test}}$ , respectively:
\[\mathcal{X}^{\text{Train}} =  \{ \bar{\mathbf{X}}_{\text{elec}}^{t:t+h}, \bar{\mathbf{X}}_{\text{elec}}^{t-w:t}, \bar{\mathbf{X}}^{t-w:t} \}_{t=w}^{N_\text{train}}
\]
We optimize the parameters of the model by minimizing the forecasting error 
\(\mathbf{E}^t_{\text{f}} = \widetilde{\mathbf{X}}_{\text{elec}}^{t:t+h} - \bar{\mathbf{X}}_{\text{elec}}^{t:t+h}\).
To learn meaningful and compact representations, we introduce an optional reconstruction error 
\(\mathbf{E}^t_{\text{r}} = \widetilde{\mathbf{X}}_{\text{elec}}^{t-w:t} - \bar{\mathbf{X}}_{\text{elec}}^{t-w:t}\) 
for regularization. The final training objective is expressed as:
\begin{equation*}
    \mathcal{L}(\mathbf{\widetilde{X}}, \mathbf{\bar{X}}) = \sum_{t=w}^{N_\text{train}} \left( \lambda_{\text{f}} \| \mathbf{E}^t_{\text{f}} \|_F^2 + \lambda_{\text{r}} \| \mathbf{E}^t_{\text{r}} \|_F^2 \right)
\end{equation*}
During the model's training process, we identify the optimal parameters:
\begin{equation}
    \theta^*, \phi^* = \arg\min_{\theta, \phi} \mathcal{L}(\mathbf{\widetilde{X}}, \mathbf{\bar{X}}; \phi, \theta)
\end{equation}

\subsection{Node Features and Graph Representation}
\label{sec:node_features}

For the forecasting problem, we consider spatially distributed sensor sites modeled as nodes (vertices) \(v \in V\) of a graph that spans the pumped storage hydropower plant.
Due to differences in raw sensor sampling rates, we use resampled time series 
based on simple moving averages, taking into account the true sensor sampling rate of the $j$-th sensor $S_j$.
This step smooths the time series:
\begin{equation}
    \bar{\mathbf{X}}[i,j] = S(\mathbf{X}[i,j]) =  \frac{1}{S_j} \sum_{\tau=1}^{S_j} s_\tau^j
\end{equation}
We model each measurement site, containing one or more sensors as an individual node. 
Each node is associated with a feature vector \(\mathbf{x}_v^{t-w:t} \in \mathbb{R}^{w \times d}, \quad \forall v \in V\), 
 containing a window $w$ of the smoothed sensor data and additional $d-1$ covariate dimensions such as a temporal encoding.
This strategy is uniformly applied to both the electrical and hydraulic components within the pumped-storage power plant environment.
Nodes are exclusively assigned to one of the sets:
\(\mathds{1}_{\text{el}}(v) = 1 \)  for electrical components or \(\mathds{1}_{\text{hyd}}(v) = 1 \) for hydraulic components, 
ensuring  \(\mathds{1}_{\text{hyd}}(v) + \mathds{1}_{\text{el}}(v) = 1\). 
As input for the subsequent model $M$, we consider the joint graph:

\[\mathcal{G}_\phi = \left(V_{\text{el}} \cup V_{\text{hyd}}, E_\phi(\bar{\mathbf{X}}[t-w:t])\right) \]

where the edges  may be learned by a parameterized function $E_\phi$.

\subsection{Attention-based Graph Learning}
\label{sec:graph_learning}

We define a trainable function that implements self-attention among the sensor nodes to infer the edges 
\(E_\phi(\mathbf{X})\) of the graph $\mathcal{G}$.
To compute the self-attention, we first map the time series 
to an embedding space \(\mathbf{E} = \text{GRU}(\bar{\mathbf{X}})\) using a gated recurrent unit (GRU).
We then proceed by computing the self-attention of the embedded time series.
For this purpose, we define a query sequence $Q$ and  a key sequence $K$ to compute the attention scores $W$: 
\begin{equation}
    Q = EW^Q, K=EW^K, W = \text{Softmax}\left( \frac{QK^T}{\sqrt{d}} \right)
\end{equation}
by linear projection with the trainable matrices \(\phi = ({W^Q, W^K})\).
Unlike in \textit{graph attention networks} (GAT) \cite{wuGraphAttentionEnabled2022}, 
which  define attention over features of a pre-existing graph,
we directly compute the symmetrically normalized graph Laplacian $L$ from the attention scores $W$ , which we convert into a symmetrical adjacency matrix
\(A = \frac{1}{2} (W + W^T)\).
We compute the Laplacian as 
\(
    L = I - D^{-\frac{1}{2}} A D^{-\frac{1}{2}}
\),
where $L$ is the Laplacian matrix, $I$ is the identity matrix, $D$ is the diagonal degree matrix of $A$, the adjacency matrix.
$L$ is then used for the graph spectral filtering in Section \ref{sec:filtering_graph_and_time}.

\subsection{Spectral-temporal Graph Neural Network}
\label{sec:stgnn}

To predict the sensor dynamics, the model processes 
the input data  $\bar{\mathbf{X}}^{t-w:t}$ 
on the learned graph $\mathcal{G}$ (obtained as introduced in Section \ref{sec:graph_learning})
by mapping the input data 
from the spatial-temporal vertex domain of the sensor signals
to a spectral latent representation.
This mapping is achieved through the sequential application of graph-spectral 
and time-spectral transformations, as introduced in Sec. \ref{sec:filtering_graph_and_time}. 
The corresponding inverse transformations are utilized to reconstruct 
the sensor signal in the spatial-temporal domain.
To address the problem of vanishing gradients and performance degradation  with increasing network depth, we introduce skip connections to compute the final forecast.
We denote the output of the residual blocks as $s_k$.
Thus, model $M$ can be expressed  with spectral filtering (\( F \)), 
and a bypass layer $f_b$ as follows in the recursive equation:
\begin{equation}
(s_k^f, s_k^b) =  
\begin{cases}
     F\left(\bar{\mathbf{X}})\mid \mathcal{G}_\phi(\bar{\mathbf{X}})\right), s^b_0 = \bar{\mathbf{X}} & k = 0\\
     F\left(\sigma(s^b_{k-1} - f_s(s^b_{k-1}))\mid \mathcal{G}_\phi(\bar{\mathbf{X}})\right),              & k > 0
\end{cases}
\end{equation}
where the final forecast is computed as \( \widetilde{\mathbf{X}}_{\text{elec}}^{t:t+h} = \Omega_f\left(\sum_{i=1}^{k} s_i^f \right)\) 
with an application-specific feedforward head function (\( \Omega \)),
and the backcast as \( \widetilde{\mathbf{X}}_{\text{elec}}^{t-w:t} = \sum_{i=1}^{k} s_i^b \).

\subsection{Graph- and Time-spectral Filtering}
\label{sec:filtering_graph_and_time}

For the spectral filtering module $F$, we use graph convolutional filtering
and the trainable \textit{Spe-Seq Cell} $S_\theta$ introduced in \cite{caoSpectralTemporalGraph2021a}.
We denote the graph Fourier transform as  \(\mathcal{GF}\),
and its inverse as \(\mathcal{IGF}\).
The $j$-th channel $y_j$ in the graph-spectral domain is therefore computed as follows:
\begin{equation}
    y_j = \mathcal{IGF}\left( \sum_i g_{\theta_{ij}}(\Lambda_i)S_\theta(\mathcal{GF}(\mathbf{X}_i)) \right).
\end{equation}
In the graph-spectral domain, we implement the parameterized filtering $g_\theta(\Lambda_i)$ on the eigenvalues $\Lambda$.
Instead of computing $y_j$ directly, we compute the Chebyshev polynomials of $L$ to efficiently approximate the graph Fourier transform 
without performing the costly eigenvalue decomposition 
of the Laplacian matrix $L = U \Lambda U^T$, where $U$ is the graph's eigenvector matrix.
We obtain the $i$-th Chebyshev polynomial $T_i(\bullet)$ with the recurrence relation:
\( T_0(x) = 1, \quad T_1(x) = x, \quad T_{k+1}(x) = 2xT_k(x) - T_{k-1}(x) \).
Thus, we implement graph-spectral filtering with the graph spectral operator $g(L)$  as follows:
\begin{equation}
g(\widetilde{L})\mathbf{X}_j \approx \sum_{n=0}^{N} c_n S_\theta(T_n(\widetilde{L})\mathbf{X}_j)
\end{equation}
where $c_n$ are the learnable parameters and $\widetilde{L} = 2L/\lambda_{\text{max}}-I_N$ is the normalized Laplacian matrix.
The \textit{Spe-Seq Cell} enhances the output of \(\mathcal{GF}\), treating it as a multivariate time-series in the graph-spectral domain. It then elevates this output  into the time-spectral domain to learn feature representations.
To achieve this,  the \textit{Spe-Seq Cell} uses the Discrete Fourier Transform (DFT)
and gated linear units (GLU) for element-wise modulation of the signal in time-spectral domain as follows:

\begin{equation}
\text{GLU}(\mathbf{x}) = \mathbf{x} \otimes \sigma(\mathbf{W}^g \mathbf{x} + \mathbf{b}^g)
\end{equation}

This approach effectively implements convolution on the multivariate time-series in the graph-spectral domain.

\section{Case Study and Experimental Setup}
\label{sec:case_study}

The dataset in this case study was obtained in collaboration with the Swiss Federal Railways (SBB).
SBB maintains a separate railway traction current network (RTN) 
that operates at a frequency of 16.66 Hz to power rolling stock across Switzerland.
The power plant operators of SBB use \textit{supervisory control and data acquisition (SCADA)} protocols to transmit sensor data 
to a centralized \textit{energy management system} (EMS). This setup allows for real-time
monitoring of assets,
ensuring timeliness and synchronicity between sensor signals,
making it a technically sound environment to evaluate the proposed methodology.

\textbf{Objective:} For this case study, our aim is to forecast the currents measured by the electrical sensor network of the PSH.
From an operator's perspective, forecasting currents is particularly compelling when dealing with rolling stock, given their highly dynamic current profile 
that is vastly different to residential power grids.
While the residential sub-grid of Zurich, the largest City of Switzerland, is subject 
to transient load changes within 15 minutes intervals of up to 35MW,
the RTN of SBB experiences load changes up to 250 MW whithin the same time interval  due to the orchestrated and periodic timetable of the Swiss railway network \cite{halderPowerDemandManagement2018}.
The importance of accurate current forecasts is additionally heightened 
because electrical components in power systems, 
like transformers and conductors, 
have thermal limitations that depend on the amount of current flowing through them.
Unlike voltage levels, current levels in the PSH dynamically react to transient loads changes.
Anomalies such as sudden increases or decreases in power demand, 
or failures in equipment, are therefore more immediately reflected in current fluctuations. 
We therefore focus on phasor current forecasting in this study.

\begin{figure}[h]
    \centering
    \includegraphics[width=1\linewidth]{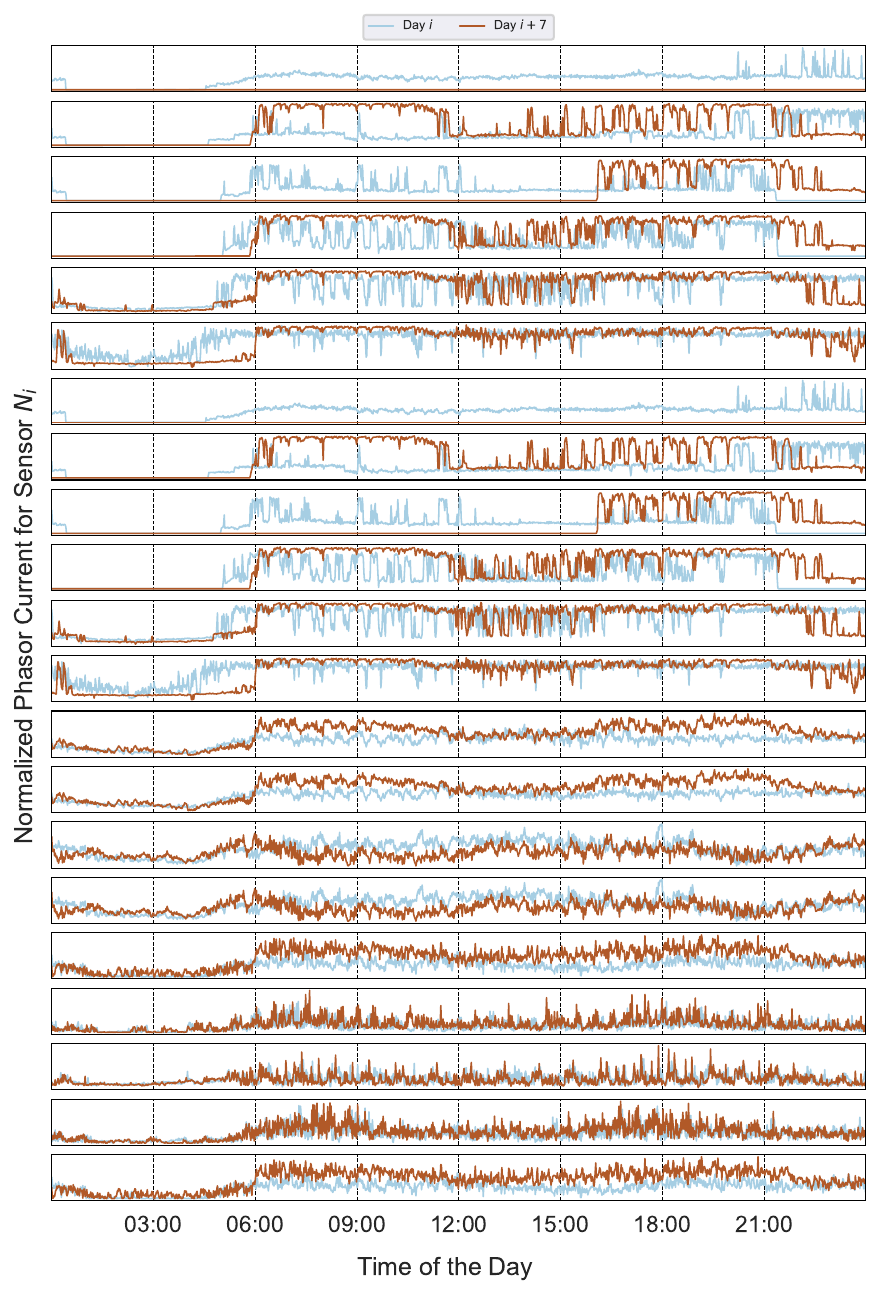}
  \caption{Segment of the dataset, displaying all 21 normalized phasor current sensors (the forecasting target of our case study), indicating the dynamic nature of the sensor measurements.  We show the same day of the week ($i$ and $i+7$) for two consecutive weeks.}
      \label{fig:zoomed_dataset}
\end{figure}

\textbf{Dataset \& Data Preparation:} We collected data spanning  four months from a PSH in Switzerland, 
consisting of readings from 58 sensors
that monitor pressures, flow rates, and lake levels of the hydraulic subsystem, as well as electrical currents from seven generating units, including connected substations 
in the electrical subsystem.
The time series are averaged to a 1-minute resolution and were collected from January to March in 2021.
We maintained the temporal ordering of training (70\% of the data), validation (15\%) and test (15\%) datasets to ensure that the validation and test indices are sequentially higher than the training indices. 
We normalize the data using feature-wise min-max scaling.
To provide  a comprehensive  understanding of the dataset, 
we show a detailed segment of the sensor data in Figure~\ref{fig:zoomed_dataset}.

\textbf{Model \& Training:}
The experiments were conducted on an NVIDIA RTX3060 
using PyTorch 2.0 and CUDA 11.8 for the development and training of the models.
The proposed model utilizes  a window size ($w$) of 24 and a horizon size of 1, meaning that it predicts the currents for the next minute.
This configuration is tailored  to the synchronized operation of the Swiss railway network, which organizes its periodic timetable in half-hourly intervals.
Our selected  model's input window takes this operational profile into account, thereby reducing the influence of the previous interval.
During model fine-tuning, we truncate the Chebychev polynomial expansion 
to $k=4$ for both the graph- and time-spectral filtering. 
We set the number of residual blocks to two and configure 
the \textit{Spe-Seq Cells} to five layers.
We adjust the negative slope of \textit{LeakyReLU} $\alpha$ to 0.2, and 
dropout to 0.5 for regularization.
We use Adam for optimization.

\begin{table}[h]
  \centering
\caption{Overview of trainable parameters in the neural network models.}

\begin{tabular}{lll}
\toprule
\textbf{Model}           & \textbf{Parameter}                           \\ \midrule
          MLP  (4-layer, el+hy)                       &  2.8 M       \\
          STF (el+hy)      &  366 K                                  \\ \midrule
          Ours (el)                &  481k                  \\
          Ours (el+hy)              & 481k                 \\ \bottomrule
\end{tabular}
\label{table:trainable_parameters}
\end{table}

\subsection{Baseline Approaches for Comparison}
\label{sec:baselines}

We compare our approach with relevant baselines for time series forecasting in the power grid domain,
such as a linear model with trend decomposition, a fully connected neural network (FNN) and a \textit{recurrent neural network},
specifically the LSTM.
 Given the  recent increase in attention towards transformer-based energy forecasting, we also consider the \textit{time series transformer} model 
(\textit{Spacetimeformer}, \cite{grigsbyLongRangeTransformersDynamic2022}, denoted as STF) 
Additionally, we include  an Attention-based GNN (A3-GCN, \cite{zhuA3TGCNAttentionTemporal2020})
that has been demonstrated to outperform the classical GCN 
on similar forecasting tasks. 
All baseline models were trained and validated 
on the same dataset and with the same input window size.
Each model was trained to reach convergence on the validation dataset.

\section{Results}

\label{sec:results}

In this section, we summarize the numerical results to 
evaluate the proposed method and compare it to the baselines.
Additionally, we study the output from the graph learning and compare it to the connectivity of the PSH asset.
Furthermore, we ablate the hydraulic information from the model input to assess its benefits.
All results are based on the dataset introduced in Section 
\ref{sec:case_study}.

In the initial step, we compare the performance of the adopted 
\textit{spectral-temporal graph neural network} 
to the baseline models introduced in Section \ref{sec:baselines} based on normalized
mean squa\-red error (NMSE).
We summarize model performances in Table \ref{tab:results}.
Our model surpasses  all conventional baselines, including 
LSTM (by 28\%) and the, in terms of parameters, much larger 
FNN (by 14\%), as evaluated by the NMSE.
A3-GCN is unable to integrate hydraulic information due to the lack of a graph. 
Spacetimeformer (STF) can learn from hydraulic signals but does not outperform our method.

\begin{table*}[htbp]
  \centering
  \caption{Average (normalized) model performance across nodes, comparing six different methods. 
  We indicate whether the PSH network diagrams were translated into a processable graph for the computation (Network Diagram)
  and epmhasize if hydraulic (Hyd.) or electric (El.) information was used for training.}
    \begin{tabular}{lccccc}
        \toprule
          \textbf{Method}    & \textbf{El.} & \textbf{Hyd.}    & \textbf{Network Diagram}      &  \textbf{Type}                  & \textbf{NMSE}                  \\
        \midrule
          Linear         & \cmark & \cmark   & \xmark        & -                    & 1.11e-1               \\
          A3-GCN         & \cmark & \xmark   & \cmark        & GCN                    & 8.74e-3               \\
          LSTM           & \cmark & \cmark   & \xmark        & RNN                    & 7.51e-3               \\
          MLP  (3-layer) & \cmark & \cmark   & \xmark        & FNN                    & 6.84e-3               \\
          MLP  (4-layer) & \cmark & \cmark   & \xmark        & FNN                    & 6.21e-3               \\ \midrule
          STF            & \cmark & \xmark   & \xmark        & Transformer            & 5.84e-3               \\
          STF            & \cmark & \cmark   & \xmark        & Transformer            & 5.83e-3               \\ \midrule
          Ours           & \cmark & \xmark   & \xmark        & Att. GCN               & \underline{5.71e-3}   \\ 
          Ours           & \cmark & \cmark   & \xmark        & Att. GCN               & \textbf{5.34e-3}      \\
        \bottomrule
    \end{tabular}
  \label{tab:results}
\end{table*}

% comparison to A3tgcn
For the evaluation of A3-GCN,
we translate the PSH’s electrical network diagram
into a processable graph, as having  apriori graph 
is a computational requirement for the method.
Surprisingly, we found that the A3-GCN is outperformed 
by the much simpler LSTM by 14\% in terms of NMSE.
This finding highlights that the intuitive approach  of applying
GNN directly to a graph derived from schematic diagrams,
does not always yield  acceptable results.
In this context, since our proposed STGNN is also based on GCN, 
the 34.7\% improvement in NMSE illustrates  that, beyond cho\-osing the right model,
finding a suitable computational graph is crucial  for processing  PSH data.
Our results provide further support 
for the observation  in~\cite{ringsquandlPowerRelationalInductive2021}
 that statistical properties 
of graphs derived from network diagrams of power grids
may be unsuitable for direct graph processing.
Graphs derived from such network diagrams significantly differ   from those typically discussed in the graph-theoretical literature, with statistical properties like 
lower clu\-ster\-ing-coefficients, 
lower node degrees, 
and higher graph diameters, 
which could explain the subpar performance of A3-GCN 
in the state forecasting task.
From a message-passing perspective, 
the specific  properties of these graphs hinder effective  message propagation unless the GNN comprises  many layers.
Unfortunaltely, this model choice, in turn, significantly  boosts 
oversmoothing, which is already a prevalent challenge in the power grid environment
due to the high similarity of the electrical sensor data.

We assess the performance of 
time-series transformers (Spacetimeformer), 
which are structurally similar to our attention-based GNN approach,
because they incorporate a self-attention layer 
across the one-\-dimensional time\-series.
How\-ever, the experiment with Spacetimeformer displays a 9.2\% reduction in performance in terms of NMSE compared to the STGNN. Additionally, we found it difficult to scale Spacetimeformer to the problem without overfitting.
Altrough showcasting respectable performance, compared to the STGNN, STF did not benefit from the additional information from hydraulic  systems
(resulting in a 0.1\% improvement).

\begin{figure}[htbp]
    \centering
    \includegraphics[width=1\linewidth]{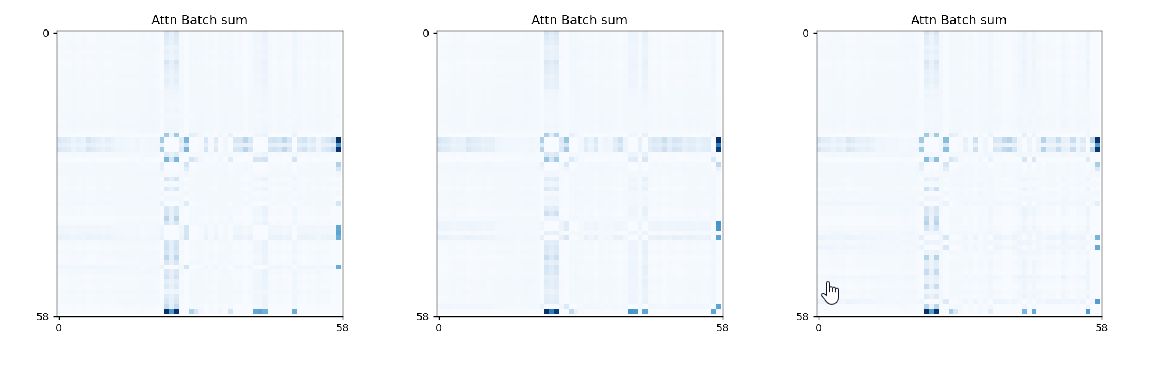}
  \caption{
  The averaged learned attention across the test set of the \textit{attention-based graph learning} module over all 58 signals from the electrical and hydraulic subsystems. We show three random seeds. The learned attention is stable for different random initializations.}
\label{fig:attention_stability}
\end{figure}

An advantage of our STGNN, compared to conventional me\-thodology 
suited for multivariate time series analysis such as LSTM, is that 
we have access to the learned graph topology.
To derive insights from the inferred graph, 
we calculate the mean attention \(a_{ij} = \frac{1}{N} \sum_{i=1}^N a_{ij} \) 
over the test data set.
We expect our method to converge to the same graph 
for randomized training initialization
when learning physical relationships between the sensors.
To verify our expectation, we visually compare the average attention
across random seeds in Figure \ref{fig:attention_stability}.
Additionally, our analysis reveals that the inferred attention graph's minimal parameterization
yields temporally stable graphs, 
accurately reflecting the situation in the PSH, 
which usually has stable topology across time.

\begin{figure*}[htbp]
    \centering
    \includegraphics[width=0.85\linewidth]{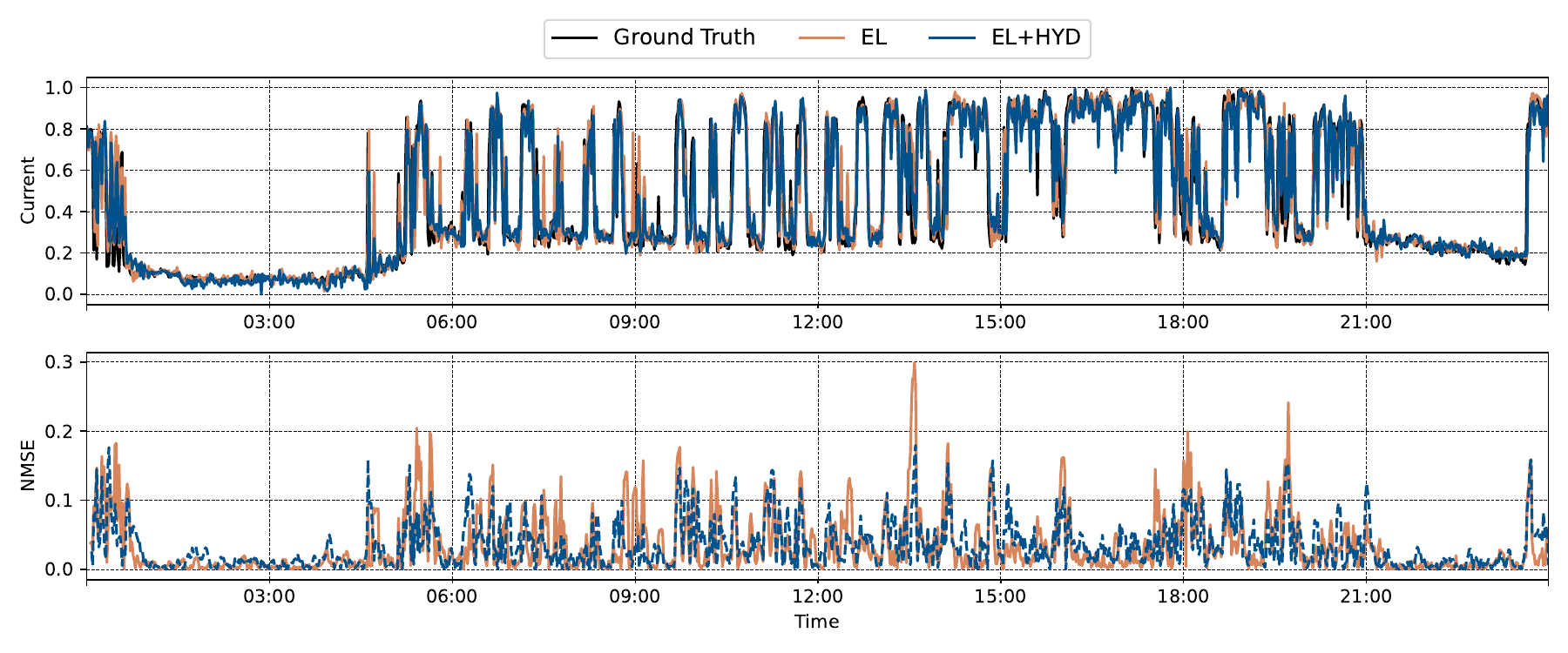}
  \caption{
    Comparison of normalized phasor current forecasts
    with (EL+HYD) and without the hydraulic information (EL) 
    for our proposed STGNN model.
    We show the forecast for single node $N_i$ ($i = 1$) across a randomly selected day including ground truth.
    In the upper Figure, we display the dynamic range of the forecast.
    In the lower Figure, we display the normalized MSE of both approaches with respect to the ground truth.
    Removing hydraulic information results in heightened discrepancies and more pronounced outliers in the predictions.
    First, we select the data based on the above criteria. Then, we normalize the selected data using min-max scaling.
  }
      \label{fig:fc_quality}
\end{figure*}

\begin{figure}[htbp]
    \centering
    \includegraphics[width=0.7\linewidth]{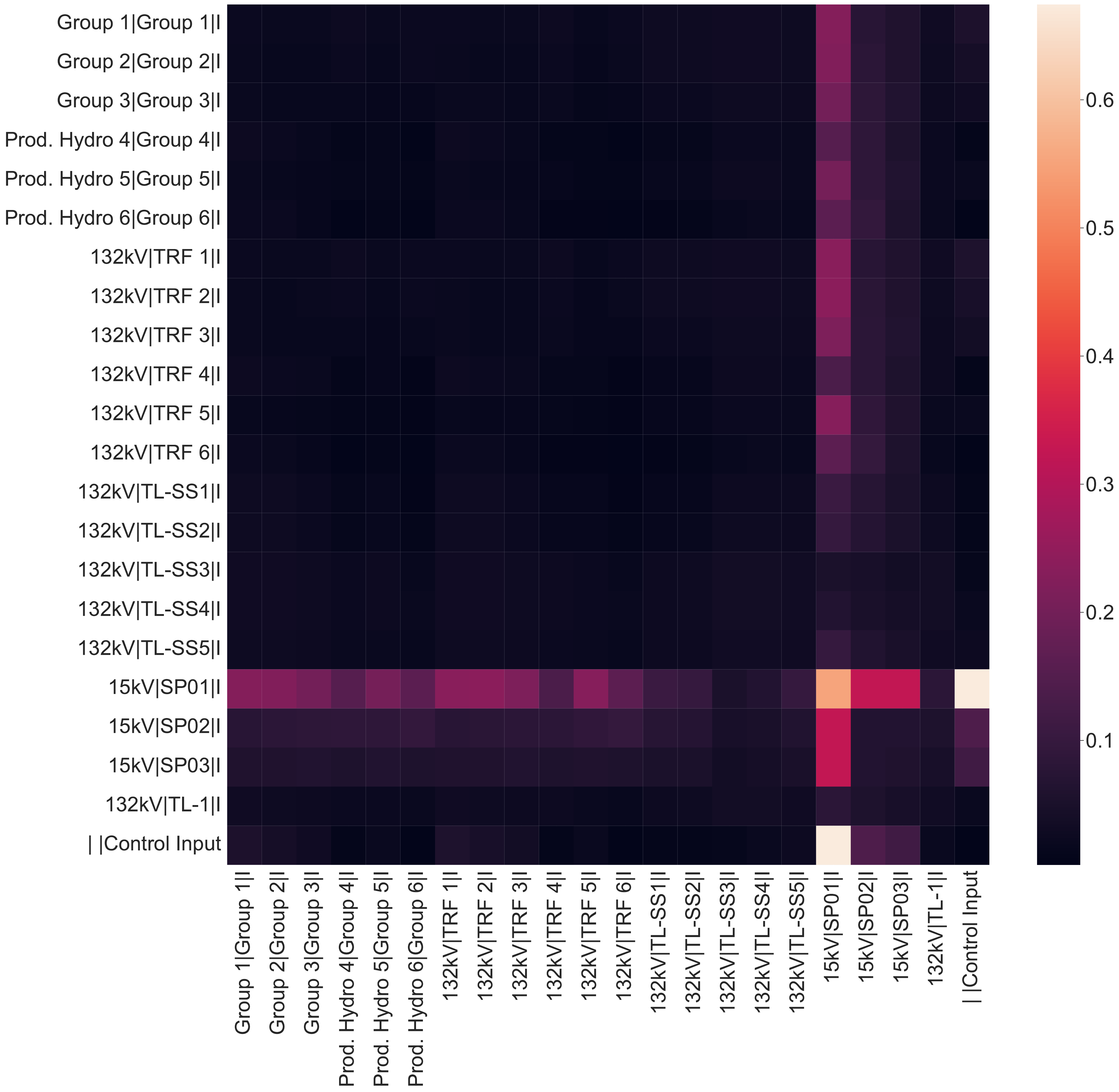}
  \caption{
  The heatmap represents the averaged learned attention by the \textit{attention-based graph learning} module across the test set as 
  for the model processing only electrial information (EL). 
  Counterintuitively, the model   focuses on  the PSH output
  (SP) when forecasting the phasor currents of the electromagnetic generators, which represent the PSH input.}
\label{fig:att_heatmap_el_el}
\end{figure}

\begin{figure*}[ht]
    \centering
    \includegraphics[width=0.6\linewidth]{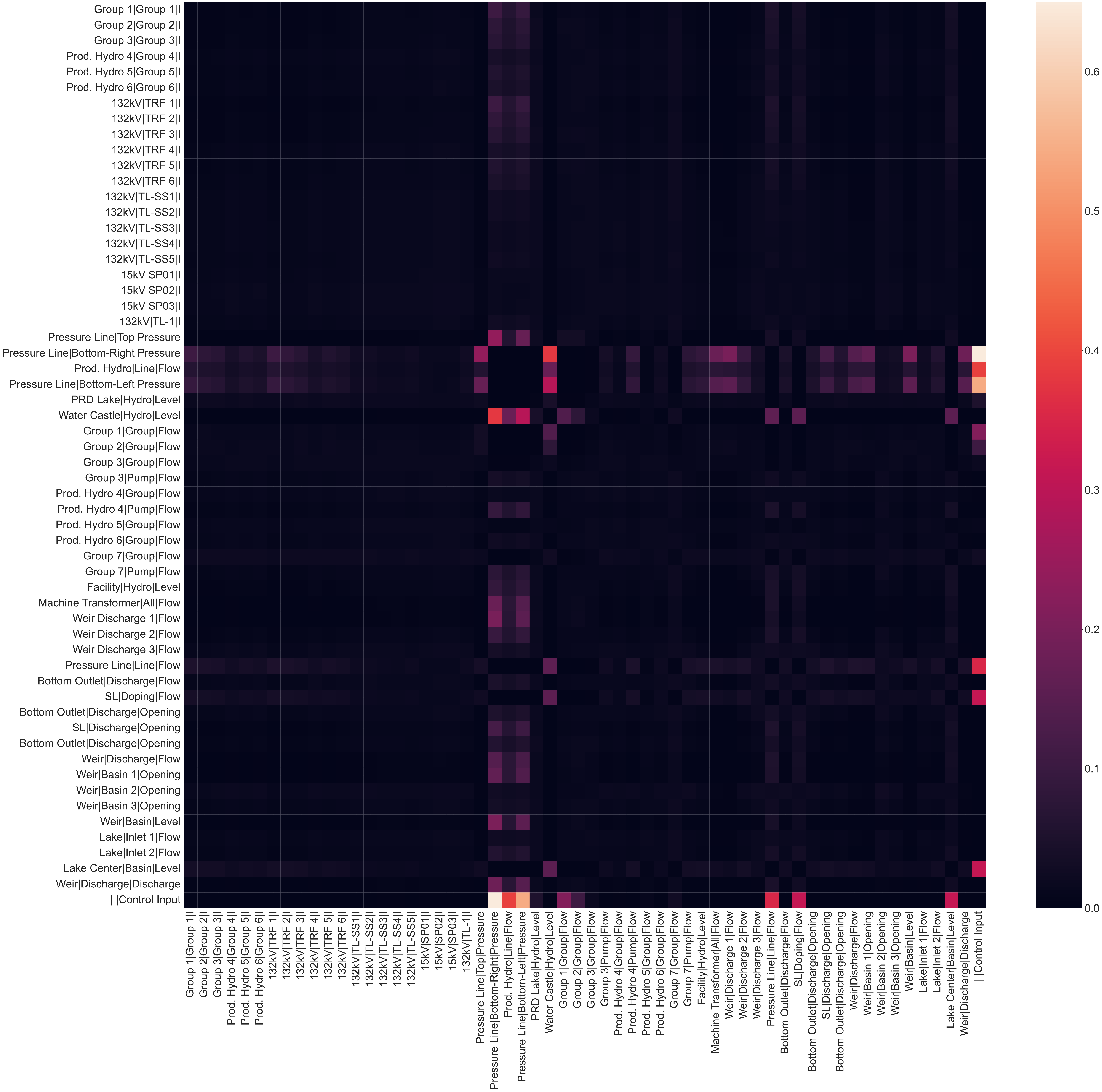}
  \caption{
  The heatmap visualizes the averaged learned attention of the \textit{attention-based graph learning} module across the test set visualized 
  for the model that processes both electrial and hydraulic information (EL+HYD). 
  Notably,  the model focuses on the hydraulic subsystem (the PSH input) 
 when  forecasting the phasor currents of the electromagnetic generators.}
\label{fig:att_heatmap_all_el}
\end{figure*}

Interestingly, the attention graph recovers casual relations given by the functioning
of the hydropower plant and therefore shows similarity to
the underlying physical network of the hydropower plant.
In Figure \ref{fig:att_heatmap_all_el}, we depict the relationships between
the water inflow (flow-rates, pressures), 
the generator units (groups and transformers, indicated by TRF), 
and the PSH outlets (substations, denoted by SP),
overall making the model's predictions more interpretable.
Leveraging this interpretability, 
we compare Figures \ref{fig:att_heatmap_el_el} and \ref{fig:att_heatmap_all_el}.
Surprisingly, our model focuses on the PSH outlets to predict the power plant 
input's phasor currents in the absence of hydraulic information from the forecast,
which could explain the more severe outliers in the forecast (Figure \ref{fig:fc_quality}).
When adding hydraulic information to the forecast, we observe that the model
makes additional use of penstock flow-rate and pressure sensor data, 
thereby improving the prediction quality.

\subsection{Ablation Study}
\begin{figure}[ht]
    \centering
    \includegraphics[width=1\linewidth]{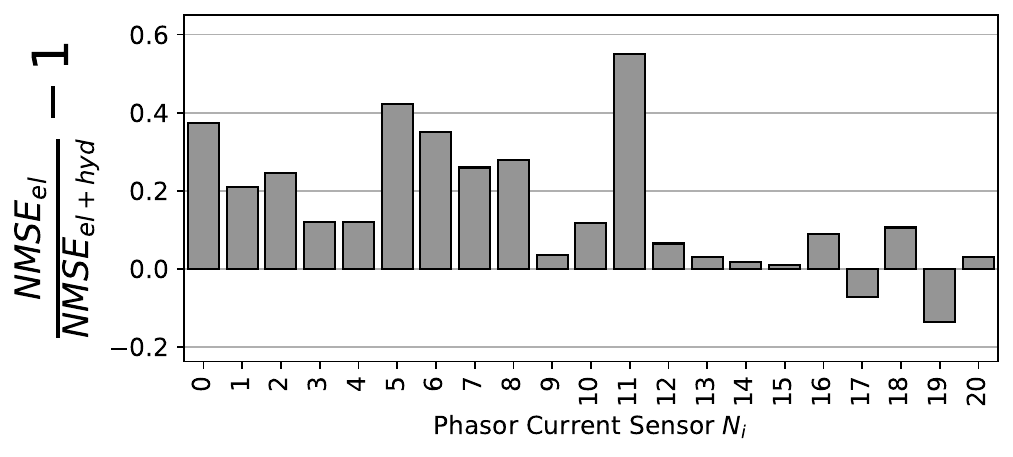}
  \caption{
      Relative improvements across the test set (normalized MSE, averaged)
      for our STGNN models with (EL+HYD) and without hydraulic information (EL).
      for each of the 21 phasor currents sensors of the electric subsystem. 
      19 out of 21 phasor current forecasts are improved by the additional hydraulic information. 
  }
\label{fig:rel_improvements}
\end{figure}

\begin{figure}[ht]
    \centering
    \includegraphics[width=1\linewidth]{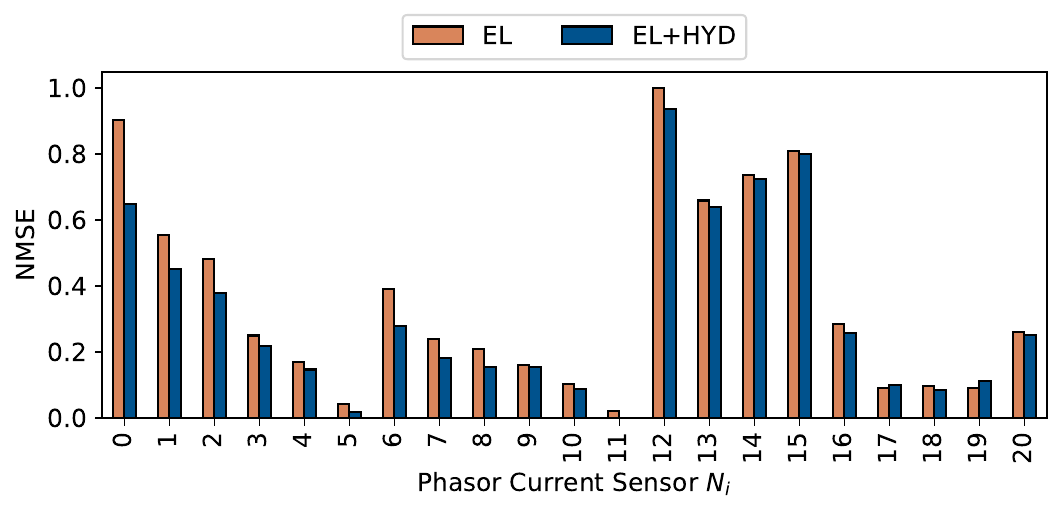}

  \caption{
      The normalized MSE averaged across the test set is displayed, comparing 
      our STGNN models with (EL+HYD) and without hydraulic information (EL).
     The performance across the 21 phasor currents sensors of the electric subsystem is shown. 
     The additional hydraulic information improves the phasor current forecasts in 19 out of the 21 cases. 
  }
\label{fig:abs_improvements}
\end{figure}

In an ablation study, we exclude the hydraulic sensor data from the forecast to validate the effectiveness of fusing electric and hydraulic sensor data for improving the electrical state forecasting. 
We find that incorporating the hydraulic subsystem leads to an 6.5\% reduction in NMSE.
The  NMSE absolute forecasting performance from the ablation experiment is included in Table \ref{tab:results}. 
Figure \ref{fig:rel_improvements} and \ref{fig:abs_improvements} evaluate the relative improvement on a per-node basis,
demonstrating that the benefits of our methodology are distributed across most sensor forecasts (nodes),
thereby ensuring that no sensor forecast experiences a major decline in performance
from including hydraulic sensor information.

\section{Conclusions}
\label{sec:conclusions}

In this paper, we demonstrate  that integrating information across the electrical and hydraulic subsystems is beneficial for state forecasting in pumped-storage hydropower plants (PSH).
Our proposed \textit{spectral-temporal graph neural network}
is the first approach to integrate information across
the PSH's subsystems by applying \textit{attention-based graph learning}, which 
effectively represents PSH states 
for short-term phasor current forecasting. 
Compared to numerical simulation, 
our me\-thod requires neither knowledge of the underlying
network and sensor connectivity graph 
nor a tedious calibration step.
Through  a real world case study,
we demonstrate that relying exclusively  on graphs derived from network diagrams for state forecasting does not always yield the best performance.
We highlight  the advantages  of learning a PSH-wide graph, complementing the critical perspective 
on network-diagram-derived graphs 
introduced in \cite{ringsquandlPowerRelationalInductive2021}.
Moreover, we show that our method remains interpretable,
unlike other deep-learning methods 
that process electrical and hydraulic data simultaneously.
Future work looks to reintegrate the underlying network diagram
while maintaining the flexibility of attention-based graph learning,
thereby harnessing the strengths of both approaches.
This could also allow for the incorporation of physics-informed losses, such as electromagnetic
generator efficiency or power flow, 
which may reduce the volume  
of training data  required.

\bibliographystyle{apacite}
\PHMbibliography{PHME2024}

\end{document}